\documentclass[letterpaper]{article} 
\usepackage{fix-cm}
\usepackage{aaai2026}  

\usepackage{times}  
\usepackage{helvet}  
\usepackage{courier}  
\usepackage[hyphens]{url}  
\usepackage{graphicx} 
\usepackage{natbib}  
\usepackage{caption} 
\usepackage{algorithm}
\usepackage{algorithmic}

\usepackage{booktabs}
\usepackage{multirow}
\usepackage{amssymb}
\usepackage{pifont}
\usepackage{amsmath}
\usepackage{newfloat}
\usepackage{listings}
\DeclareCaptionStyle{ruled}{labelfont=normalfont,labelsep=colon,strut=off} 
\floatstyle{ruled}
\newfloat{listing}{tb}{lst}{}
\floatname{listing}{Listing}
\nocopyright 

\title{PAD-F: Prior-Aware Debiasing Framework for Long-Tailed X-ray Prohibited Item Detection }
\author {
    Haoyu Wang\textsuperscript{\rm 1}\equalcontrib,
    Renshuai Tao\textsuperscript{\rm 1}\equalcontrib,
    Wei Wang\textsuperscript{\rm 1}, \thanks{Corresponding author.}
    Yunchao Wei\textsuperscript{\rm 1}
}
\affiliations {
Beijing Jiaotong University\\
{\fontsize{8.5pt}{\baselineskip}\selectfont \tt \{ wanghy23,rstao, wei.wang, yunchao.wei\}@bjtu.edu.cn}

}

\usepackage{bibentry}

\begin{document}

\maketitle

\begin{abstract}
Detecting prohibited items in X-ray security imagery is a challenging yet crucial task. With the rapid advancement of deep learning, object detection algorithms have been widely applied in this area. However, the distribution of object classes in real-world prohibited item detection scenarios often exhibits a distinct long-tailed distribution. Due to the unique principles of X-ray imaging, conventional methods for long-tailed object detection are often ineffective in this domain.
To tackle these challenges, we introduce the Prior-Aware Debiasing Framework (PAD-F), a novel approach that employs a two-pronged strategy leveraging both material and co-occurrence priors. At the data level, our Explicit Material-Aware Augmentation (EMAA) component generates numerous challenging training samples for tail classes. It achieves this through a placement strategy guided by material-specific absorption rates and a gradient-based Poisson blending technique. At the feature level, the Implicit Co-occurrence Aggregator (ICA) acts as a plug-in module that enhances features for ambiguous objects by implicitly learning and aggregating statistical co-occurrence relationships within the image.
Extensive experiments on the HiXray and PIDray datasets demonstrate that PAD-F greatly boosts the performance of multiple popular detectors. It achieves an absolute improvement of up to +17.2\% in AP50 for tail classes and comprehensively outperforms existing state-of-the-art methods. Our work provides an effective and versatile solution to the critical problem of long-tailed detection in X-ray security.
\end{abstract}

\section{Introduction}
\label{sec:intro}

As crowd density increases in public transportation hubs \cite{carvalho2017understanding,wagner2020evaluation}, ensuring public safety through effective security inspections has become more critical than ever. X-ray scanners \cite{withers2021x}, widely deployed to inspect luggage and produce complex imagery, play a crucial role in these inspections\cite{5cd9402ee1cd8e8e2dd1558f,5b1643398fbcbf6e5a9ba699,5c62c410f56def97988aeeda}. However, the accuracy of identifying prohibited items can be compromised due to the limitations of manual inspection, particularly under conditions of prolonged concentration, which may lead to missed detection of prohibited items and potential security breaches. Therefore, there is a pressing need for an accurate and automated detection to support these safety efforts. Advances in deep learning \cite{lecun2015deep,khan2019deep}, particularly CNNs, offer promising solutions by framing AI inspection as a detection task \cite{zou2023object,zhao2019object,5390b72e20f70186a0f21766} in the computer vision community~\cite{tan2020efficientdet,5cda948de1cd8ecf46bb4b94}.

\begin{figure}[!t]
  \centering
   \includegraphics[width=\linewidth]{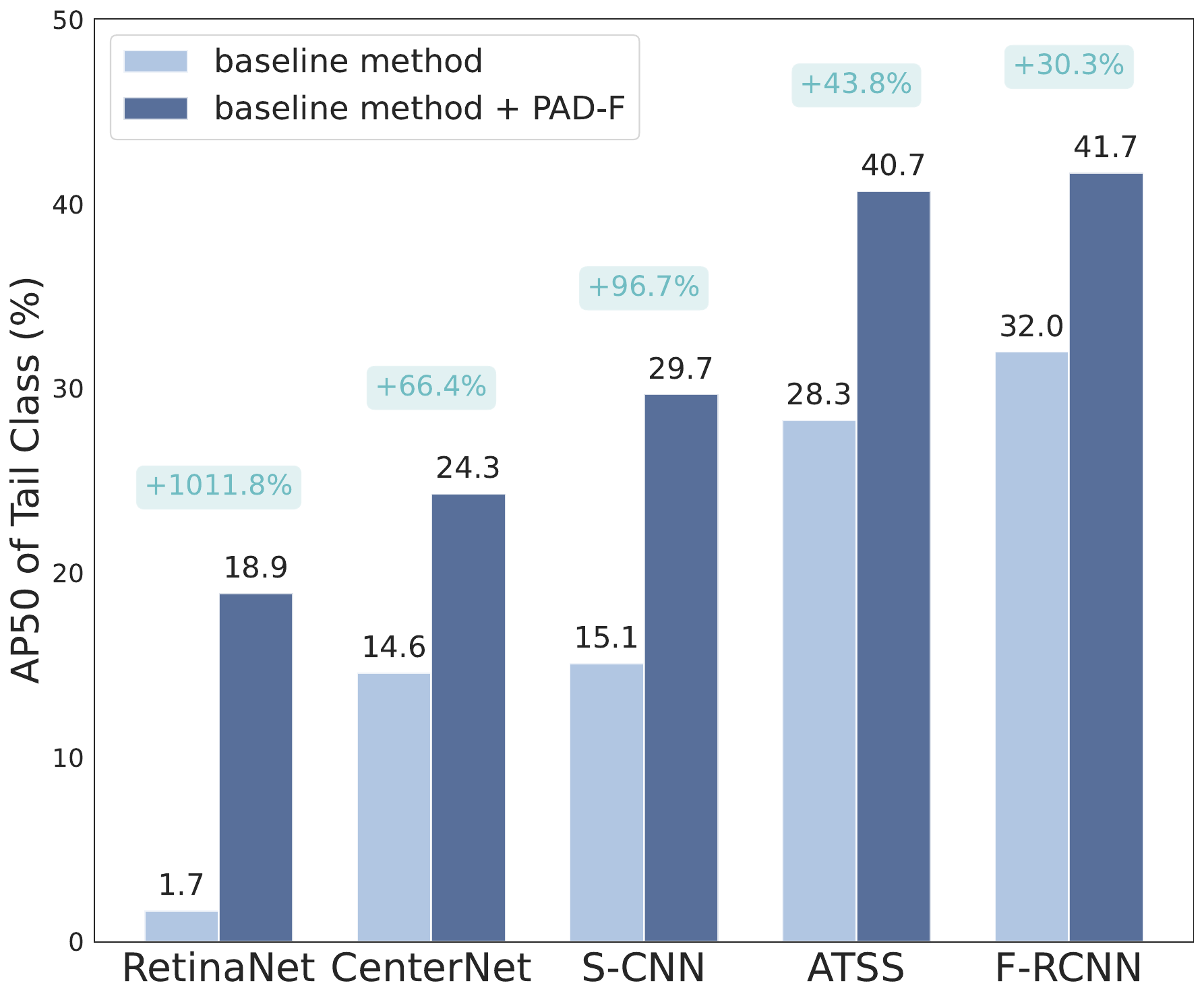}
   \caption{Improvement of tail object performance.}
   \label{fig:intuition}
\end{figure}

Conventional CNN-based approaches \cite{xiao2018deep, fu2018refinet} have demonstrated promising results in X-ray object detection, leveraging broad advancements in the field, showing significant improvements in identifying a wide range of prohibited items. However, in real-world scenarios, the frequency of different categories varies, leading to a long-tail distribution in prohibited item detection data, as seen in datasets like HiXray \cite{Hixray}. Although many effective algorithms exist for long-tailed object detection in natural images \cite{zhao2024logit}, they generally perform poorly on X-ray data. To date, research dedicated to long-tailed prohibited item detection in X-ray security images is notably limited.

The primary cause of this predicament stems from the unique image formation properties of X-rays. Unlike in natural scenes, where imaging relies on light reflection, an X-ray image is formed based on the differential attenuation of X-rays as they pass through objects of varying material compositions. This principle leads to two inherent complexities: intricate inter-object overlap and a scarcity of low-level texture features. Furthermore, the apparent visual properties of an object, such as its color and opacity, are not intrinsic but context-dependent, changing based on the other objects with which it is superimposed.
\begin{figure}[!t]
  \centering
   \includegraphics[width=0.9\linewidth]{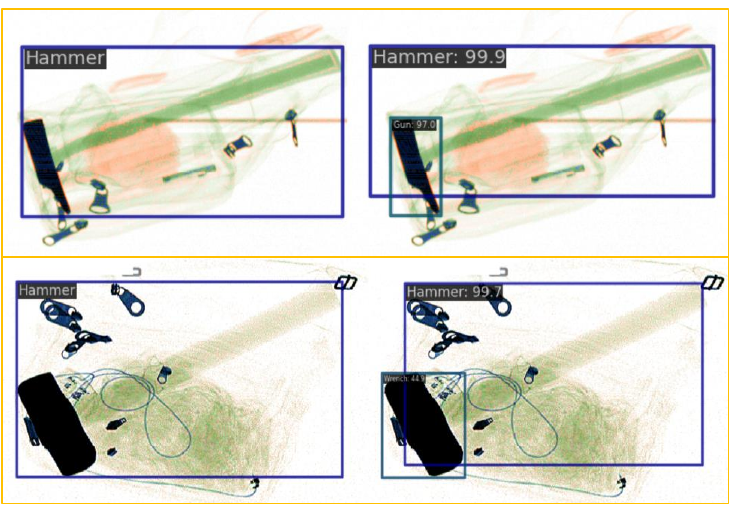}
   \caption{Experimental visualization. The left side shows the ground truth, while the right displays the predictions. We observe the difficulty in distinguishing between categories. 
   }
   \label{fig:mis-cls}
\end{figure}
This implies a fundamental distinction from natural images. \textbf{In a typical image, an unoccluded object (i.e., at the top of the visual layer) possesses a complete and stable set of identifiable attributes—such as the metallic sheen of a tool or the distinct pattern of animal fur. This principle, however, does not hold for X-ray imagery.}
These unique characteristics manifest as two significant challenges for long-tail prohibited item detection:
(1)For long-tail problems, augmenting tail-class data via methods like copy-and-paste is a direct and highly effective strategy. However, due to the context-dependent appearance of objects in X-ray scans, such simple augmentation techniques are rendered invalid, as they cannot realistically simulate the complex interplay of overlapping materials.
(2)The loss of fine-grained superficial details results in impoverished semantic representations of objects. This can lead to a high rate of misclassification. For instance, as illustrated in Fig. \ref{fig:mis-cls} (left: ground truth; right: baseline detector output), the head of a hammer, characterized by its solid metal composition, monolithic color, and simple geometry, is misidentified by the model as a wrench, another class defined by similar primitive features.

\begin{figure*}[!h]
  \centering
   \includegraphics[width=1\linewidth]{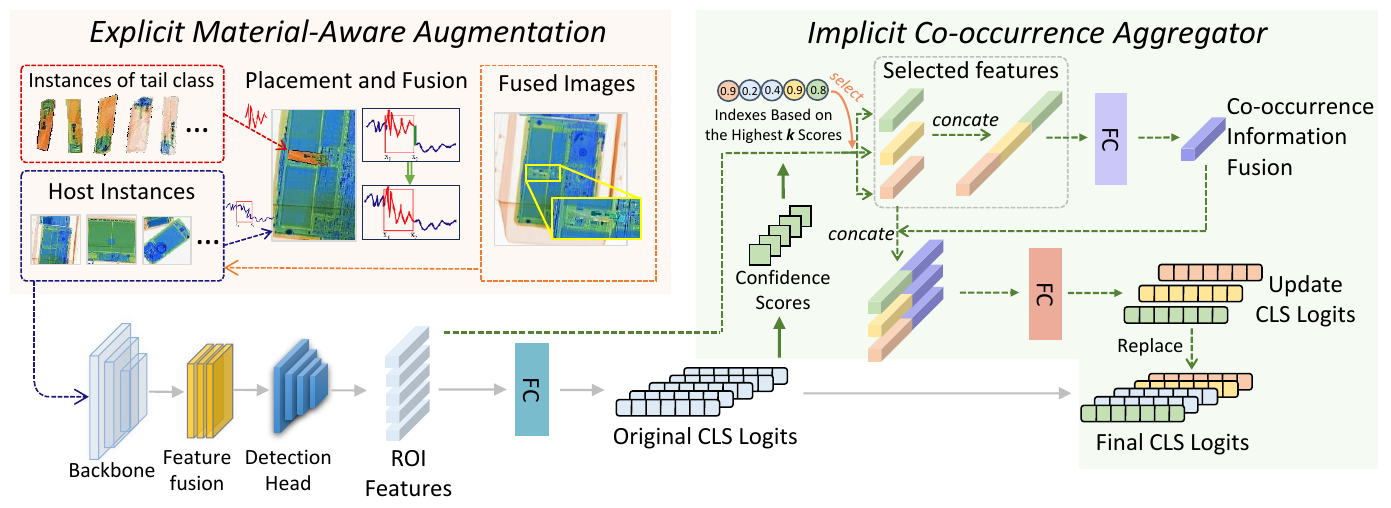}
   \caption{The framework overview of PAD-F. The framework employs a synergistic strategy where two components work in concert. First, the EMAA component enriches the training data by prior-guided augmentation to debias the data distribution. Second, the ICA component, embedded within the detector head, refines proposal features by incorporating co-occurrence patterns learned from the most salient objects. This dual approach tackles both long-tail distribution and feature ambiguity.}
   \label{fig-method}
\end{figure*}

To tackle this dual challenge of statistical data bias and imaging-induced ambiguity, we propose the Prior-Aware Debiasing Framework (PAD-F). Our framework employs a two-pronged approach: it explicitly leverages material priors at the data level while implicitly learning statistical co-occurrence priors at the feature level. Concretely, we introduce two core components: (1) The Explicit Material-Aware Augmentation (EMAA) module directly confronts the problem of physical camouflage. Unlike generic augmentation methods, EMAA simulates physically plausible overlaps based on material properties, generating high-quality, challenging samples that compel the model to learn to distinguish camouflaged tail-class items from dominant head-class objects. (2) The Implicit Co-occurrence Aggregator (ICA) is designed to compensate for the feature-weakness of rare objects. This plug-in module implicitly learns relationships between co-occurring objects within the scene, allowing it to enhance the features of ambiguous items when they appear alongside their common counterparts, thereby providing crucial co-occurrence evidence for the final detection. Figure \ref{fig:intuition} shows the impact of our PAD-F module on several popular baseline detectors. As can be seen, our method yields substantial performance gains for the tail classes. In summary, our main contributions are as follows:
\begin{itemize}
\item We propose EMAA, a novel data augmentation strategy guided by material priors. It makes augmentation effective for the X-ray long-tail problem by realistically simulating object camouflage and overlap, thus overcoming a key limitation of conventional methods.

\item  We design ICA, a lightweight and effective plug-in module that learns statistical co-occurrence patterns to enhance the representations of ambiguous objects, thereby improving the performance of detectors.

\item We conduct extensive experiments on challenging detection benchmarks. The results demonstrate that our proposed PAD-F achieves state-of-the-art performance and brings notable improvements to various baseline detectors, proving its effectiveness and generalizability.

\end{itemize}




\section{Related Work}\label{Section:relatedwork}
\subsection{X-ray prohibited item detection}
X-ray imaging offers a powerful ability in many tasks, such as medical image analysis \cite{zhang2021window, chaudhary2019diagnosis,lu2019towards,Wang_2025_CVPR,chen2024cariesxrays} and security inspection \cite{5b1643398fbcbf6e5a9ba699,miao2019sixray, huang2019modeling,ji2021prohibited,gao2024114xray}. As a matter of fact, obtaining X-ray images is difficult, few studies touch on security inspection in computer vision due to the lack of specialized high-quality datasets.
Recently, multiple datasets have been made available for research\cite{miao2019sixray,WeiOccluded2020,Hixray,EDS,FSOD,PIDray,CLCXray}\cite{WeiOccluded2020}. These datasets facilitate the evaluation of tasks such as image classification, object detection, cross-domain detection, and few-shot detection. Furthermore, numerous studies have extensively investigated image recognition and reconstruction in the X-ray domain.\cite{Cai_2024_CVPR} proposed a Structure-Aware method for 3D reconstruction from sparse-view X-ray images.\cite{duan2023rwsc} designed a region fusion model for X-ray images, tailored to the luminance and saturation characteristics of X-ray images.\cite{yang2024dual} attempts to solve the problems of domain discrepancy and label inconsistency encountered when consolidating multiple datasets for X-ray prohibited item detection.

\subsection{Long-tail Vision Tasks}
Long-tail visual recognition addresses the imbalance between common (head) and rare (tail) classes in real-world datasets. Several approaches have been proposed to handle the challenge of learning effective representations for tail classes\cite{li2024feature,zhang2021bag,hsieh2021droploss}. Class rebalancing strategies like \textbf{re-sampling} and \textbf{re-weighting} aim to mitigate bias towards head classes. For example, \cite{cui2019class} introduced an effective number of samples to create class-balanced loss weights, improving performance for underrepresented classes. \cite{copypaste} efficiently augments object detection datasets through a simple copy-paste operation on instances. Similar instance-level data augmentation strategies are also presented in \cite{zhang2017mixup} and \cite{devries2017improved}.

Recent work, such as Logit Normalization (LogN) by \cite{zhao2024logit}, self-calibrates classified logits, similar to batch normalization, to address long-tail recognition. \cite{zhang2023reconciling} proposed ROG, a method that reconciles object-level and global-level objectives in long-tail detection.\cite{kang2020decoupling} introduced Decoupled Training, which first learns a feature extractor using cross-entropy loss and then fine-tunes it with a rebalanced classifier. \cite{li2019gradient} addresses the class imbalance problem by focusing on the gradient density of samples. Additionally, contrastive learning methods like \cite{tang2020long} align feature distributions to ensure consistency between head and tail classes, more robust to imbalance.\cite{wang2021seesaw} mitigates the issue of head classes excessively suppressing tail classes in long-tailed data distributions through a dynamic re-balancing strategy.

\section{Methodology}
\label{sec:methodology}



\subsection{Overall Framework}
\label{sec:overall_framework}

To systematically address the dual challenges of statistical scarcity and imaging-induced ambiguity, we introduce the Prior-Aware Debiasing Framework (PAD-F).
As illustrated in Figure~\ref{fig-method}, our framework enhances a standard object detector on two prior-aware fronts:

To explicitly leverage the prior knowledge of how different material properties impact X-ray imaging, we introduce the \textbf{Explicit Material-Aware Augmentation (EMAA)} strategy. Unlike naive copy-paste methods, EMAA analyzes the material properties of a host image to place and fuse tail-class items into regions where they are most likely to be camouflaged. This process generates a rich set of challenging training samples that compel the model to learn the critical, often subtle, patterns of tail-class items, even when embedded within complex backgrounds.

To compensate for the scarcity of reliable local features caused by X-ray imaging principles, we embed the \textbf{Implicit Co-occurrence Aggregator (ICA)} module within the detector's architecture. Strategically positioned between the RoI feature extractor and the final classification heads, the ICA module learns to aggregate statistical co-occurrence relationships among object proposals. This process enhances the feature representations of tail-category items, which are often ambiguous when viewed in isolation, by providing crucial contextual evidence for the final decision.

The two components of PAD-F form a synergistic strategy: EMAA focuses on manipulating the input data distribution and exposing the model to more challenging scenarios, while ICA enhances the model's internal inference mechanism to better integrate ambiguous features. The entire framework can be seamlessly integrated with existing detectors and trained end-to-end, effectively boosting long-tail object detection performance in complex security scenarios without introducing significant inference overhead.



\subsection{Explicit Material-Aware Augmentation}\label{data-level}

The goal of our Explicit Material-Aware Augmentation (EMAA) module is to generate training samples that are both physically plausible and highly challenging, directly targeting the unique difficulties of X-ray object detection. EMAA achieves this through a two-stage process: a novel placement strategy followed by a smooth fusion strategy.

\subsubsection{\textbf{Material-Aware Placement Strategy}}
Augmenting the data for tail classes is an effective strategy for addressing the long-tail distribution problem. However, a drawback of conventional augmentation methods like Copy-Paste \cite{ghiasi2021simple} is their stochastic placement policy. They typically insert foreground objects at random locations within a host image. This often results in objects being placed in simple, low-clutter areas, creating trivial training samples that fail to teach the model how to handle cases of camouflage and heavy occlusion, which is not suit for X-ray imagery.

To overcome this limitation, the first key component of our EMAA is a material-aware placement strategy. The core of this strategy is to maximize the learning challenge by purposefully fusing objects with dissimilar material attributes. In X-ray imaging, an object's attenuation is primarily determined by its material. Generally, metallic objects exhibit higher attenuation and appear darker, while non-metallic objects have lower attenuation and appear lighter.

As shown in Figure \ref{fig-method}, our algorithm pairs a tail-class instance with a host object from a head-category based on their contrasting X-ray attenuation properties. For example, a low-attenuation tail object is strategically placed onto a high-attenuation, structurally complex region of a host object. This synthesis of material-aware challenging samples compels the model to learn to identify faint object patterns amidst a visually dominant and complex background, thereby enhancing its performance on tail-class objects.
\begin{figure}[!t]
  \centering
   \includegraphics[width=\linewidth]{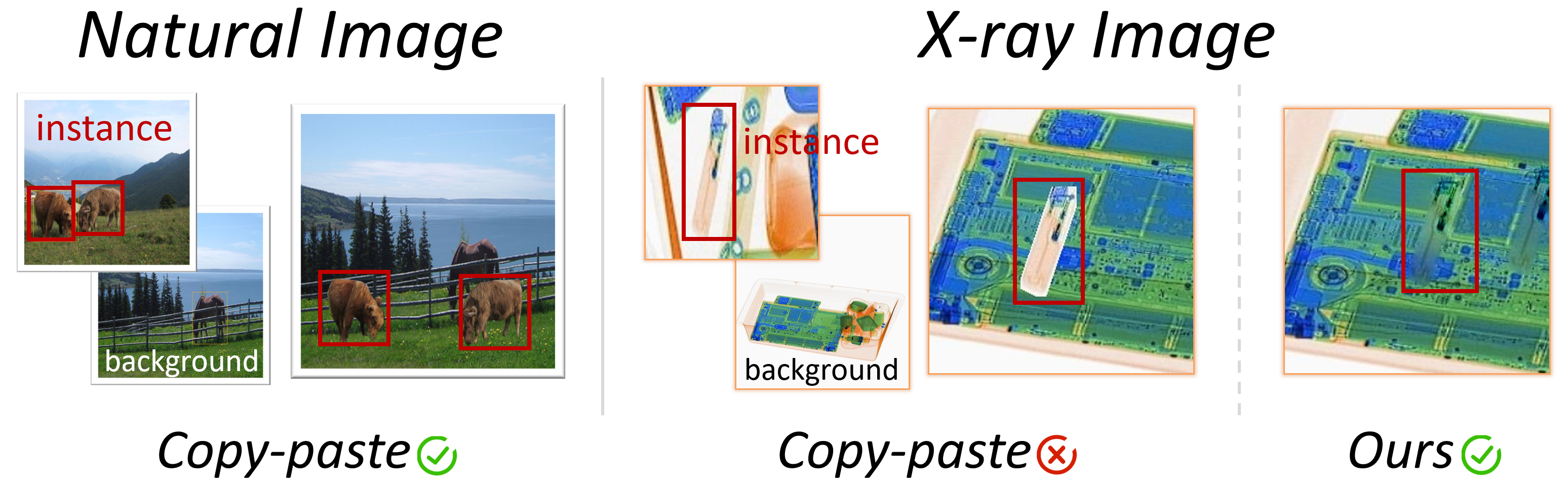}
   \caption{Comparison to natural image data augmentation.}
   \label{fig:copy-paste}
\end{figure}
\subsubsection{\textbf{Fusion Strategy}}

\label{sec:fusion_strategy}
Once a material-aware placement is determined, the next challenge is to fuse the tail-class instance into the host image smoothly. 
Figure \ref{fig:copy-paste} illustrates the performance gap of the standard copy-paste data augmentation technique when applied to natural and X-ray domains, comparing it against our method. For natural images, the visual appearance of an object instance remains unchanged when pasted into new scenes, making simple copy-paste sufficient for generating new training samples. Conversely, in the X-ray domain, this naive pasting process introduces abrupt artifacts and a contextual gap between the instance and the background, caused by perspective imaging in the X-ray scenario. To generate better training samples, we employed the Poisson blending method based on gradient fields.



In general, the core idea is to preserve the source object's gradient field while forcing its pixel intensities to conform to the target area. This ensures a smooth transition without sacrificing the internal details of the object being pasted.

Let $\Omega$ be the region of the tail object to be pasted, with a boundary $\partial\Omega$. Let $g$ be the source (tail) image and $f^*$ be the target (host) image. The goal is to compute the new pixel values $f$ within the region $\Omega$ of the final fused image. This is formulated as a variational problem to find a function $f$ that minimizes the following objective:
\begin{equation}
\label{eq:variational}
\min_{f} \iint_{\Omega} |\nabla f - \mathbf{v}|^2 \,dx\,dy \quad \text{with} \quad f|_{\partial\Omega} = f^*|_{\partial\Omega}
\end{equation}
Here, $\mathbf{v} = \nabla g$ is the guidance vector field, which is the gradient field of the source image $g$. The boundary condition $f|_{\partial\Omega} = f^*|_{\partial\Omega}$ ensures that the pixel values at the boundary of the pasted region seamlessly match the background.

This minimization problem is equivalent to finding the solution of the Poisson equation with Dirichlet boundary conditions:
\begin{equation}
\label{eq:poisson}
\Delta f = \text{div}(\mathbf{v}) \quad \text{over} \quad \Omega, \quad \text{with} \quad f|_{\partial\Omega} = f^*|_{\partial\Omega}
\end{equation}
where $\Delta$ is the Laplacian operator and $\text{div}$ is the divergence operator. In essence, this method intelligently modifies the pixel intensities of the tail object so that its texture seamlessly integrates with the lighting and texture of the local background in the host image, creating a highly realistic and artifact-free composite. The benefits of this approach will be quantitatively validated in our experiments section.

\subsection{ Implicit Co-occurrence Aggregator }\label{model-level}

While EMAA enhances the model's ability to handle physical camouflage at the data level, tail-category objects often suffer from intrinsically weak or ambiguous features. For example, a small, rare prohibited item may be visually similar to a part another object or a cluttered background texture. In such cases, the object's identity cannot be reliably determined from its appearance alone. However, valuable clues can often be found in the surrounding co-occurrence objects.

To systematically exploit these contextual cues, we introduce the Implicit Co-occurrence Aggregator (ICA), a lightweight and plug-in module designed to enhance object features by aggregating statistical co-occurrence priors. As shown in Figure 2, the ICA operates on the set of region proposals within an image, allowing each object's feature representation to be refined by aggregating information from its surrounding context. This process enables the model to make more informed predictions, especially for the items whose individual appearance is ambiguous.

In this module, we enhance the detection process by leveraging the contextual relationships between objects and surroundings in the image. The first step is to select the top $k$ proposals based on their confidence scores. Let the set of initial proposals be $\{p_1, p_2, \dots, p_N\}$, with corresponding confidence scores $\{s_1, s_2, \dots, s_N\}$. We then select the top $k$ proposals, denoted by $P_{\text{top-k}} = \{p_{i_1}, p_{i_2}, \dots, p_{i_k}\}$, where the confidence scores satisfy $s_{i_1} \geq s_{i_2} \geq \dots \geq s_{i_k}$. These proposals are assumed to contain the most likely prohibited items and will serve as the basis for further enhancement.
\begin{algorithm}[!t]
	\caption{Training Procedure of PAD-F}
	\label{alg:PAD_F_training}
	\begin{algorithmic}[1] 
		\REQUIRE Input images $I_k$, bounding boxes $B_k$, class labels $C_k$, for all $k \in [1, K]$.
		\ENSURE Trained model $\mathbf{M}$.
		
		\STATE \textbf{X-ray-Specific Augmentation:}
\FOR{each tail class $c \in C_{\text{tail}}$}
    \FOR{each instance $b_i^k \in B_k$ of class $c$}
        \STATE Crop $I_{c, i} = I_k[b_i^k]$
        
        \STATE Select a host image $I_{\text{bg}}$ containing an object $b_{\text{host}}$
        \STATE \quad \textbf{where} Attenuation($b_{\text{host}}$) contrasts with Attenuation($I_{c,i}$)
        \STATE Determine a new location $b_{\text{new}}$ over the region of $b_{\text{host}}$
        
        \STATE Generate $I_{\text{new}} = \text{ImageFusion}(I_{\text{bg}}, I_{c, i}, b_{\text{new}})$
        
        \STATE Replace $(I_{\text{new}}, B_{\text{new}}, C_{\text{new}})$ to dataset $D'$
    \ENDFOR
\ENDFOR
		
		\STATE \textbf{Contextual Feature Integration:}
		\STATE Sample Batches from the $D'$;
		\STATE Generate proposals $\{p_1, p_2, \dots, p_N\}$ using RPN, with scores $\{s_1, s_2, \dots, s_N\}$;
		\STATE Select $P_{\text{top-k}} = \{p_{i_1}, p_{i_2}, \dots, p_{i_k}\}$ based on scores;
		\STATE Concatenate $F_{\text{concat}} = \text{Concat}(f_{i_1}, f_{i_2}, \dots, f_{i_k})$;
		\STATE Compute fusion $F_{\text{fusion}} = \sigma(\mathbf{W}_1 F_{\text{concat}} + \mathbf{b}_1)$;
		\FOR{each feature vector $f_{i_j}$ in $P_{\text{top-k}}$}
		    \STATE Compute $f_{i_j}^{\text{aug}} = \text{Concat}(f_{i_j}, F_{\text{fusion}})$;
		    \STATE Update feature $f_{i_j}^{\text{new}} = \sigma(\mathbf{W}_2 f_{i_j}^{\text{aug}} + \mathbf{b}_2)$;
		\ENDFOR
		\STATE Replace original logits with $\text{Logits}_{\text{new}} = \text{ClassificationHead}(f_{i_j}^{\text{new}})$;
		\STATE Perform backpropagation and update parameters;
		\STATE Output the trained model $\mathbf{M}$.
	\end{algorithmic}
\end{algorithm}
For each of these top-$k$ proposals, we extract their feature vectors, denoted as $f_{i_j}$ for $j = 1, 2, \dots, k$. To capture the relationships among these top proposals, we concatenate their feature vectors into a single vector:

\begin{equation}
F_{\text{concat}} = \text{Concat}(f_{i_1}, f_{i_2}, \dots, f_{i_k}),
\end{equation}

This concatenated feature vector $F_{\text{concat}}$ contains information about the dependencies between the top-$k$ proposals, reflecting how they interact in the context of the image. We then apply a fully connected layer to process $F_{\text{concat}}$ and produce a relational fusion feature, $F_{\text{fusion}}$, as follows:

\begin{equation}
F_{\text{fusion}} = \sigma(\mathbf{W}_1 F_{\text{concat}} + \mathbf{b}_1),
\end{equation}
where $\sigma$ is an activation function, such as ReLU, and $\mathbf{W}_1$ and $\mathbf{b}_1$ are the weights and bias of the fully connected layer. The relational feature $F_{\text{fusion}}$ captures the higher-order interactions between the selected proposals, encoding the contextual relationships that are crucial for distinguishing between visually similar categories.

To further integrate this relational information, we concatenate the fused relational feature $F_{\text{fusion}}$ with each of the original proposal features $f_{i_j}$, forming an augmented feature vector for each proposal. We then apply another fully connected layer to each augmented feature vector to produce the final updated feature vector $f_{i_j}^{\text{new}}$. This process can be formulated as follows: 

\begin{equation}
f_{i_j}^{\text{aug}} = \text{Concat}(f_{i_j}, F_{\text{fusion}}),
\end{equation}

\begin{equation}
f_{i_j}^{\text{new}} = \sigma(\mathbf{W}_2 f_{i_j}^{\text{aug}} + \mathbf{b}_2),
\end{equation}
where $\mathbf{W}_2$ and $\mathbf{b}_2$ are the weights and bias of the second fully connected layer. This updated feature vector $f_{i_j}^{\text{new}}$ incorporates both the original features of the proposals and the relational information from other proposals, enabling the model to make more informed classification decisions.

Finally, the updated features $f_{i_j}^{\text{new}}$ replace the features before classification and regression heads. The new classification logits are computed from these enhanced features:

\begin{equation}
\text{Logits}_{\text{new}} = \text{ClassificationHead}(f_{i_j}^{\text{new}}),
\end{equation}

By incorporating the spatial and semantic dependencies between objects, this module enhances the model's ability to distinguish between similar categories and improve detection performance, particularly for tail categories in challenging environments with occlusions or cluttered backgrounds. This contextual enhancement helps the model recognize prohibited items more accurately, even when they are visually similar to other objects or surrounded by complex scenes.

\subsection{Overall Training Process}\label{training_process}
Algorithm \ref{alg:PAD_F_training} outlines the entire training procedure for PAD-F. First, the EMAA module employs a material-aware placement strategy before applying image fusion to create challenging training data. Second, the ICA enhances proposal features by aggregating co-occurrence information from high-confidence objects. By integrating these two synergistic modules, PAD-F effectively addresses the long-tail distribution problem in X-ray security inspection, delivering superior performance on underrepresented tail categories.

	
	

\section{Experiments}\label{sec:experiments}


\begin{table*}[!t]
	\begin{center}
        \footnotesize
		\setlength{\tabcolsep}{0.54mm}{
			\small
			\begin{tabular}{c|c|ccccccc|c|c|cccccccccccc}
				\midrule
				\multirow{2}{*}{\centering Method}& \multicolumn{9}{c|}{HiXray Dataset} & \multicolumn{13}{c}{PIDray Dataset}\\
                \cmidrule(lr{8pt}){2-10} \cmidrule(lr{8pt}){11-23}
				& \textbf{AP\textsubscript{50}}& PO1 & PO2 & WA & LA & MP & TA & CO & NL (\textbf{Tail}) & \textbf{AP\textsubscript{50}} & BA & PL & HA & PO & SC & WR & GU & BU & SP & HA & KN & LI  \\
				\midrule
				F-RCNN &84.2&95.0&93.4&91.9&98.0&97.0&95.7&71.0&32.0&76.0&83.7&96.0&88.5&60.7&91.2&96.7&47.6&71.9&67.3&98.7&42.1&67.8 \\
				+ \textbf{ours}&\textbf{85.2}&96.1&94.4&92.4&98.0&97.0&94.1&68.0&\textbf{41.7}$_{{\uparrow 9.7\%}}$&\textbf{77.6}&87.7&96.4&90.0&63.2&92.5&97.0&48.2&72.6&72.8&98.7&42.8&69.6   \\
				\midrule
                S-RCNN&80.8&94.2&92.4&90.3&98.2&97.0&95.1&63.6&15.1&71.8&84.4&95.0&86.7&62.3&86.4&96.8&37.1&66.0&48.6&98.5&33.9&65.7\\
                + \textbf{ours}&\textbf{83.2}&94.8&93.6&91.2&97.9&97.3&94.5&66.5&\textbf{29.7}$_{{\uparrow 14.6\%}}$&\textbf{72.4}&85.0&94.5&83.3&63.5&86.3&96.3&38.5&66.8&56.0&98.6&34.3&66.5\\
				\midrule
				RetinaNet&79.0&94.0&93.2&90.8&98.1&97.3&95.4&61.8&1.7&69.0&77.6&93.1&82.8&48.9&83.7&91.6&38.4&64.2&58.2&98.0&28.0&63.5   \\
				+ \textbf{ours}& \textbf{81.8}&94.8&93.8&91.0&97.9&97.5&95.1&65.6&\textbf{18.9}$_{{\uparrow \textbf{17.2}\%}}$&\textbf{70.5}&77.8&93.8&84.9&48.8&84.8&92.6&47.3&63.6&58.1&98.0&31.9&64.4   \\
				\midrule
				CenterNet&82.4&95.6&94.3&90.7&98.1&97.8&95.3&73.1&14.6&74.0&86.5&96.8&90.3&56.9&90.6&95.6&35.3&68.7&67.9&98.5&35.3&65.7\\
                + \textbf{ours}&\textbf{83.7}&95.5&94.5&90.9&98.0&97.9&95.4&73.2&\textbf{24.3}$_{{\uparrow 9.7\%}}$&\textbf{74.3}&87.0&96.9&91.4&54.7&90.3&97.0&36.9&72.8&66.7&98.4&33.3&65.8   \\
				\midrule
                ATSS&83.5&95.2&93.6&91.8&98.2&97.4&95.4&68.0&28.3&74.6&84.1&96.3&91.2&56.5&90.9&96.4&39.8&70.6&71.4&98.5&33.0&66.7\\
                + \textbf{ours}&\textbf{85.6}&95.7&94.2&93.1&98.1&97.4&95.4&70.6&\textbf{40.7}$_{{\uparrow 12.4\%}}$&\textbf{76.2}&87.6&96.9&90.0&58.2&90.0&96.6&49.3&72.1&74.7&98.0&32.8&67.8\\
				\bottomrule
			\end{tabular}
		}
	\end{center}
	\caption{Comparison of performance between baselines and PAD-F-enhanced (+ours) versions on the HiXray and PIDray datasets. The proposed PAD-F consistently improves AP\textsubscript{50} of the overall categories, especially for the tail ones in HiXray.
	}
	\label{table-traditional}
\end{table*}

\subsection{Experimental Setup}\label{sec:experimental_settings}

\textbf{Tasks and Datasets}. To ensure fair comparisons with previous detection methods, we evaluate our model on two \textbf{large-scale} X-ray datasets, HiXray \cite{Hixray} and PIDray \cite{PIDray}. HiXray is derived from real-world scenarios with a distinctly long-tail distribution, making it ideal for evaluating model performance on naturally imbalanced data. Although PIDray samples are synthetically generated and balanced, this dataset offers a comprehensive evaluation framework for X-ray imaging tasks and is suitable for testing the classification capabilities of the proposed PAD-F.

\textbf{Evaluation Metrics}. Following standard object detection practices, we use Average Precision 50(AP\textsubscript{50}) to assess the model’s category-specific performance and overall effectiveness. AP\textsubscript{50} is calculated at Intersection over Union (IoU) thresholds 0.5, providing a robust measure of the model’s overall performance on detecting items.

\textbf{Implementation Details}. All the experiments were conducted using the mmdetection open-source toolbox\footnote{\url{https://github.com/open-mmlab/mmdetection}} \cite{chen2019mmdetection} to ensure consistency and comparability. The parameter settings for all models followed the default configurations of mmdetection. Experiments were conducted on four NVIDIA GeForce RTX 4090 GPUs.

\subsection{Comparison with Various Detection Baselines}\label{sec:model_integration}
To determine the performance and generalizability of PAD-F, we integrated it with multiple popular object detection frameworks, including F-RCNN (Faster RCNN \cite{ren2015faster}), S-RCNN (Sparse RCNN \cite{sun2021sparse}), RetinaNet \cite{ross2017focal}, CenterNet \cite{duan2019centernet}, and ATSS \cite{zhang2020bridging}. 

\textbf{Performance on HiXray Dataset}.The experimental results in Table \ref{table-traditional} compellingly demonstrate the effectiveness of the PAD-F framework. The overall AP\textsubscript{50} sees consistent improvement across all baselines; the biggest gains are observed in the NL (Tail) class. The standout result is with the RetinaNet model, where the performance on the NL class surged from a mere 1.7\% to 18.9\%, achieving an absolute improvement of 17.2\%. This dramatic enhancement highlights the framework's ability to effectively learn features for extremely rare objects. Furthermore, the method shows strong generalizability by substantially boosting other models as well. For instance, S-RCNN's performance on the tail class jumped by 14.6\%, and ATSS saw a 12.4\% increase. These results confirm that PAD-F provides a robust and widely applicable solution for mitigating the long-tail problem in X-ray object detection.


\textbf{Performance on PIDray Dataset}. The PAD-F also demonstrated strong performance gains on the PIDray dataset. Faster RCNN, for instance, saw an increase in AP50 from 76.0\% to 77.6\%. Notable improvements include:
(1) The AP of class GU improved from 47.6\% to 48.2\%. This improvement, while relatively small, demonstrates PAD-F's ability to provide more robust features for objects that may be occluded or embedded within a complex context. Guns, in particular, often have parts that overlap with other objects, and the improvement suggests better discrimination for such features.
(2) The AP of class SP rose from 67.3\% to 72.8\%. This increase indicates that PAD-F's augmentations made the model better at recognizing screwdrivers even when partially occluded or in cluttered environments, which is a common scenario in X-ray imagery.  For instance, using S-RCNN with PAD-F on the PIDray dataset resulted in an increase in AP for GU from 37.1\% to 38.5\%.




\subsection{Comparison with Long-Tail Detection Methods}\label{compare-longtail}
We compared PAD-F with several SOTA long-tail task methods in Table \ref{table-lt}. The methods considered in this comparison are LogN \cite{zhao2024logit}, ROG \cite{zhang2023reconciling}, Seasaw \cite{wang2021seesaw}, GHM \cite{li2019gradient}, and Focal \cite{ross2017focal}. These methods are popular for addressing the challenges posed by class imbalances, especially for the tail category.

Our PAD-F achieves the highest overall AP\textsubscript{50} of 85.2, outperforming all competing methods, including the next-best ROG (84.7) and Seasaw (83.8). The primary advantage of our framework is its superior performance on the tail categories. Notably, for the most challenging tail class, NL, PAD-F scores 41.7, a substantial improvement over all other approaches, such as ROG (35.2) and Seasaw (35.5). This result validates PAD-F's effectiveness in addressing the severe class imbalance inherent in X-ray security imagery.

\subsection{Comparison with Data Augmentation Methods}\label{data-aug-methods}
In this section, we compare the performance of PAD-F with several popular data augmentation methods, including Copy-paste \cite{ghiasi2021simple}, Mixup \cite{zhang2017mixup}, and Cutout \cite{devries2017improved}. These methods are widely used for enhancing the performance of object detection models, especially in the context of imbalanced datasets. The results are illustrated in Tab. \ref{table-augmentation-methods}.

 While these conventional methods perform well on head categories, they falter on tail classes. Our PAD-F achieves the highest overall $AP_{50}$ of \textbf{85.6\%}, surpassing all competitors like Copy-paste (84.4\%) and Cutout (83.9\%). The primary advantage of our method is evident in the most challenging tail class, NL, where PAD-F achieves a score of \textbf{40.7\%}. This represents a substantial improvement of over 14 absolute points compared to the best-performing conventional method, Copy-paste (26.4\%), demonstrating the clear superiority of our PAD-F for the X-ray long-tail problem.
\begin{table}[!t]
	\begin{center}
		\setlength{\tabcolsep}{0.7mm}
		{
			\small
			\begin{tabular}{c|c|ccccccc|cc} 
				\toprule
                \multirow{2}{*}{\centering Method}& \multirow{2}{*}{\centering \textbf{AP\textsubscript{50}}}&\multicolumn{8}{c}{\textbf{AP\textsubscript{50} across various categories}} \\
               \cmidrule(lr{9pt}){3-10}&& PO1 & PO2 & WA & LA & MP & TA & CO & NL\\
				\midrule

                LogN & 83.1 & 95.2 & 93.6 & 92.0&98.0&96.7&95.1&62.8&31.3 \\
ROG & 84.7 &95.8 & 94.0 & 93.1  & 98.2 &97.5&96.3&67.3&35.2\\
\midrule
                Seasaw  &83.8&95.4&94.3&92.3&98.1&97.0&88.3&69.5&35.5   \\
                GHM &80.4&93.8&90.4&88.7&97.3&96.5&92.4&53.7&30.1  \\
                Focal&82.2&94.7&93.2&89.5&97.5&97.6&91.5&63.0&30.2   \\
                \midrule
                PAD-F (\textbf{ours})&\textbf{85.2}&96.1&94.4&92.4&98.0&97.0&94.1&68.0&\textbf{41.7}   \\
				\bottomrule
			\end{tabular}
		}
	\end{center}
	\caption{Comparison with popular long-tail detection methods on the HiXray dataset. 
    (base model: Faster-RCNN)}
	\label{table-lt}
\end{table}
\begin{table}[!t]
	\begin{center}
		\setlength{\tabcolsep}{0.7mm}
		{
			\small
			\begin{tabular}{c|c|ccccccc|cc} 
				\toprule
                \multirow{2}{*}{\centering Method}& \multirow{2}{*}{\centering \textbf{AP\textsubscript{50}}}&\multicolumn{8}{c}{\textbf{AP\textsubscript{50} across various categories}} \\
               \cmidrule(lr{9pt}){3-10}&& PO1 & PO2 & WA & LA & MP & TA & CO & NL\\
				\midrule
Copy-paste & 84.4&96.3&95.7&94.1&98.0&98.2&96.4&69.7&26.4 \\
Mixup & 80.8&94.4&92.8&89.3&97.9&96.9&94.2&64.8&16.3 \\
Cutout &83.9&95.5&93.9&91.6&98.2&97.5&96.2&73.2&25.2\\
\midrule
 PAD-F (\textbf{ours})&\textbf{85.6}&95.7&94.2&93.1&98.1&97.4&95.4&70.6&\textbf{40.7}\\

\bottomrule
\end{tabular}}
	\end{center}
	\caption{Comparison with popular object data augmentation methods for detection on HiXray (base model: ATSS).}
	\label{table-augmentation-methods}
\end{table}

\subsection{Ablation Studies}\label{sec:ablation_study}
We conducted ablation studies to determine the contributions of EMAA and ICA individually, as well as the impact of the hyperparameter \( k \) of the EMAA.
\subsubsection{Effectiveness of Each Module}\label{sec:module_effectiveness}

\begin{table}[!t]
	\begin{center}
		\setlength{\tabcolsep}{1mm}
		{
			\small
			\begin{tabular}{lc|c|cccccccc} 
                \toprule

                \multicolumn{2}{c|}{\textbf{Setting}}& \multirow{2}{*}{\centering \textbf{AP\textsubscript{50}}}&\multicolumn{8}{c}{\textbf{AP\textsubscript{50} across various categories}} \\
                \cmidrule(lr{7pt}){1-2}  \cmidrule(lr{9pt}){4-11}
				 E & I&& PO1 & PO2 & WA & LA & MP & TA & CO & NL\\
				\midrule
			    \ding{55} & \ding{55} &84.2&95.0&93.4&91.9&98.0&97.0&95.7&71.0&32.0  \\
			
                 \ding{55} & \checkmark & 84.6&95.3&94.5&92.5&98.0&97.0&94.8&68.0&36.8    \\
				
                 \checkmark & \ding{55} &85.0&95.3&93.9&92.3&97.9&97.7&94.3&67.4&41.2   \\
				
                 \checkmark & \checkmark &\textbf{85.2}&96.1&94.4&92.4&98.0&97.0&94.1&68.0&\textbf{41.7}  \\
				\bottomrule
			\end{tabular}
		}
	\end{center}
\caption{Ablation study on individual contributions of EMAA (E) and ICA(I). The table demonstrates how each component contributes to the overall improvement.}
\label{table-eBFPN}  
\end{table}
The ablation results of each module are provided in Tab. \ref{table-eBFPN}. We analyzed the effects of enabling EMAA (E), ICA(I), and both.

We analyze the individual contributions of the EMAA (E) and ICA(I) in Table~\ref{table-eBFPN}. The baseline model (without EMAA or ICA) achieves an overall $AP_{50}$ of 84.2\% and 32.0\% on the tail class NL. Enabling only the EMAA (E) provides the most influential boost, raising the overall $AP_{50}$ to 85.0\% and, crucially, the NL class AP to 41.2\%. In contrast, using the ICA(I) alone offers a more modest gain, with an overall $AP_{50}$ of 84.6\% and an NL AP of 36.8\%. The full PAD-F framework, with both modules enabled, achieves the best performance, reaching a peak overall $AP_{50}$ of \textbf{85.2\%} and the highest tail-class AP of \textbf{41.7\%} on NL. This demonstrates a clear synergistic effect, where both components are vital to achieving optimal performance.

\begin{figure}[!t]
  \centering
   \includegraphics[width=0.95\linewidth]{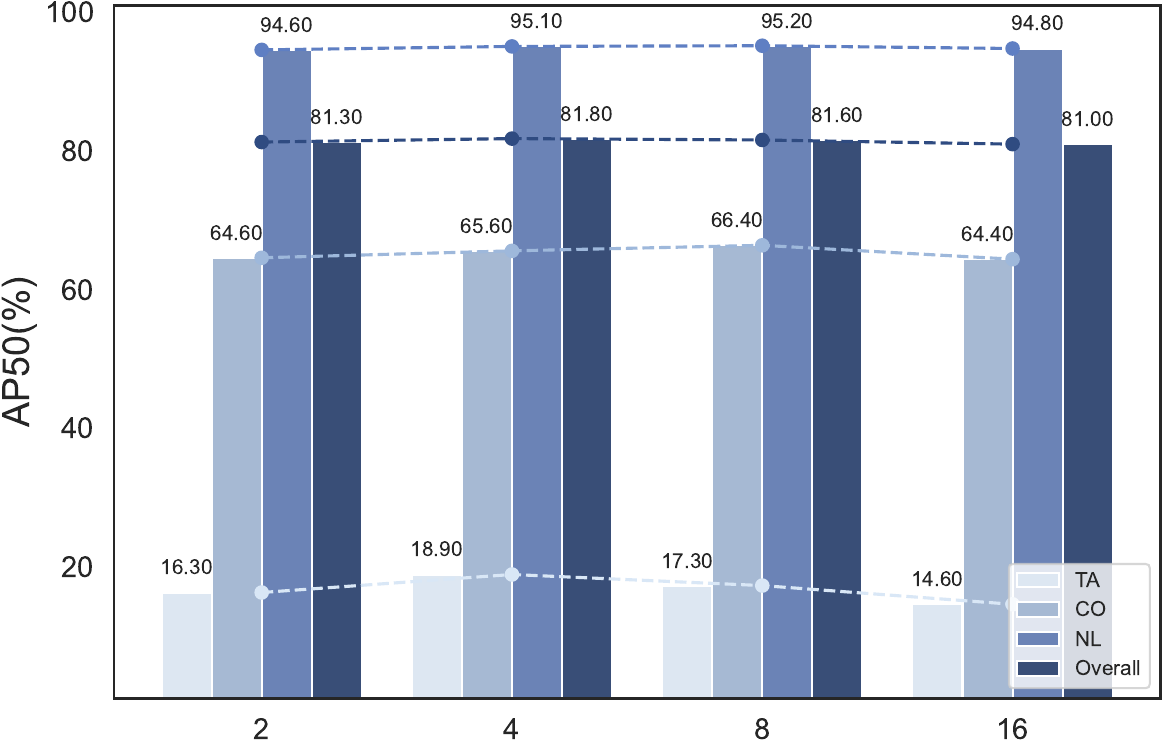}
   \caption{Performance impact of the $k$ on the ICA. Results show optimal performance when $k$ = 4.}
   \label{fig:pare-k}
\end{figure}
\subsubsection{Impact of Hyperparameters}
We analyze the impact of the hyperparameter $k$, which controls the number of context proposals in the ICA module, on RetinaNet. As shown in Figure~\ref{fig:pare-k}, performance peaks with an $AP_{50}$ of 81.8\% when $k=4$. The results show that using too few context proposals ($k=2$) leads to insufficient relational learning, while using too many ($k=8, 16$) introduces noise that degrades performance, particularly on tail classes. This study validates that a moderate amount of contextual information is optimal, and we thus set $k=4$ in all our experiments.

\section{Conclusion}\label{Section:conclusion}
This paper addresses the challenges in X-ray security inspection posed by long-tail distribution and imaging principles. To this end, we have successfully designed and validated the Prior-Aware Debiasing Framework (PAD-F), an innovative dual-level enhancement framework based on material and co-occurrence priors. 
Our PAD-F not only possesses excellent versatility in improving various baseline detectors but also exhibits superior performance on underrepresented tail classes, greatly outperforming state-of-the-art techniques.
For future work, we plan to investigate how to generalize the proposed data augmentation method to more challenging X-ray object detection tasks. Furthermore, we will research more interpretable and scalable co-occurrence relationship modeling methods, aiming to make greater contributions toward ensuring public safety.

\bibliography{aaai2026}

\begin{thebibliography}{53}
\providecommand{\natexlab}[1]{#1}

\bibitem[{Akcay et~al.(2018)Akcay, Kundegorski, Willcocks, and Breckon}]{5b1643398fbcbf6e5a9ba699}
Akcay, S.; Kundegorski, M.~E.; Willcocks, C.~G.; and Breckon, T.~P. 2018.
\newblock Using Deep Convolutional Neural Network Architectures for Object Classification and Detection Within X-Ray Baggage Security Imagery.
\newblock 13: 2203--2215.

\bibitem[{Cai et~al.(2024)Cai, Wang, Yuille, Zhou, and Wang}]{Cai_2024_CVPR}
Cai, Y.; Wang, J.; Yuille, A.; Zhou, Z.; and Wang, A. 2024.
\newblock Structure-Aware Sparse-View X-ray 3D Reconstruction.
\newblock In \emph{Proceedings of the IEEE/CVF Conference on Computer Vision and Pattern Recognition (CVPR)}, 11174--11183.

\bibitem[{Carvalho, Marques, and Costeira(2017)}]{carvalho2017understanding}
Carvalho, J.; Marques, M.; and Costeira, J.~P. 2017.
\newblock Understanding people flow in transportation hubs.
\newblock \emph{IEEE Transactions on Intelligent Transportation Systems}, 19(10): 3282--3291.

\bibitem[{Chaudhary, Hazra, and Chaudhary(2019)}]{chaudhary2019diagnosis}
Chaudhary, A.; Hazra, A.; and Chaudhary, P. 2019.
\newblock Diagnosis of Chest Diseases in X-Ray images using Deep Convolutional Neural Network.
\newblock In \emph{2019 10th International Conference on Computing, Communication and Networking Technologies (ICCCNT)}, 1--6. IEEE.

\bibitem[{Chen et~al.(2024)Chen, Fu, Liu, Pan, Lu, and Zhang}]{chen2024cariesxrays}
Chen, B.; Fu, S.; Liu, Y.; Pan, J.; Lu, G.; and Zhang, Z. 2024.
\newblock CariesXrays: Enhancing caries detection in hospital-scale panoramic dental X-rays via feature pyramid contrastive learning.
\newblock In \emph{Proceedings of the AAAI Conference on Artificial Intelligence}, volume~38, 21940--21948.

\bibitem[{Chen et~al.(2019)Chen, Wang, Pang, Cao, Xiong, Li, Sun, Feng, Liu, Xu et~al.}]{chen2019mmdetection}
Chen, K.; Wang, J.; Pang, J.; Cao, Y.; Xiong, Y.; Li, X.; Sun, S.; Feng, W.; Liu, Z.; Xu, J.; et~al. 2019.
\newblock MMDetection: Open mmlab detection toolbox and benchmark.
\newblock \emph{arXiv preprint arXiv:1906.07155}.

\bibitem[{Cui et~al.(2019)Cui, Jia, Lin, Song, and Belongie}]{cui2019class}
Cui, Y.; Jia, M.; Lin, T.-Y.; Song, Y.; and Belongie, S. 2019.
\newblock Class-balanced loss based on effective number of samples.
\newblock In \emph{Proceedings of the IEEE/CVF conference on computer vision and pattern recognition}, 9268--9277.

\bibitem[{DeVries and Taylor(2017)}]{devries2017improved}
DeVries, T.; and Taylor, G.~W. 2017.
\newblock Improved regularization of convolutional neural networks with cutout.
\newblock \emph{arXiv preprint arXiv:1708.04552}.

\bibitem[{Duan et~al.(2019)Duan, Bai, Xie, Qi, Huang, and Tian}]{duan2019centernet}
Duan, K.; Bai, S.; Xie, L.; Qi, H.; Huang, Q.; and Tian, Q. 2019.
\newblock Centernet: Keypoint triplets for object detection.
\newblock In \emph{Proceedings of the IEEE/CVF international conference on computer vision}, 6569--6578.

\bibitem[{Duan et~al.(2023)Duan, Wu, Mao, Yin, Xiong, and Li}]{duan2023rwsc}
Duan, L.; Wu, M.; Mao, L.; Yin, J.; Xiong, J.; and Li, X. 2023.
\newblock Rwsc-fusion: Region-wise style-controlled fusion network for the prohibited x-ray security image synthesis.
\newblock In \emph{Proceedings of the IEEE/CVF conference on computer vision and pattern recognition}, 22398--22407.

\bibitem[{Fu, Zhao, and Gu(2018)}]{fu2018refinet}
Fu, K.; Zhao, Q.; and Gu, I. Y.-H. 2018.
\newblock Refinet: a deep segmentation assisted refinement network for salient object detection.
\newblock \emph{IEEE Transactions on Multimedia}, 21(2): 457--469.

\bibitem[{Gao et~al.(2024)Gao, Guan, Huang, Li, Liao, Huang, Zheng, Lin, Li, Tang et~al.}]{gao2024114xray}
Gao, H.; Guan, Z.; Huang, Y.; Li, X.; Liao, H.; Huang, B.; Zheng, H.; Lin, R.; Li, L.; Tang, H.; et~al. 2024.
\newblock 114Xray: A Large-Scale X-Ray Security Detection Benchmark and Aware Enhance Network for Real-World Prohibited Item Inspection in Baggage.
\newblock In \emph{Chinese Conference on Pattern Recognition and Computer Vision (PRCV)}, 246--260. Springer.

\bibitem[{Ghiasi et~al.(2021{\natexlab{a}})Ghiasi, Cui, Srinivas, Qian, Lin, Cubuk, Le, and Zoph}]{copypaste}
Ghiasi, G.; Cui, Y.; Srinivas, A.; Qian, R.; Lin, T.-Y.; Cubuk, E.~D.; Le, Q.~V.; and Zoph, B. 2021{\natexlab{a}}.
\newblock Simple copy-paste is a strong data augmentation method for instance segmentation.
\newblock In \emph{Proceedings of the IEEE/CVF conference on computer vision and pattern recognition}, 2918--2928.

\bibitem[{Ghiasi et~al.(2021{\natexlab{b}})Ghiasi, Cui, Srinivas, Qian, Lin, Cubuk, Le, and Zoph}]{ghiasi2021simple}
Ghiasi, G.; Cui, Y.; Srinivas, A.; Qian, R.; Lin, T.-Y.; Cubuk, E.~D.; Le, Q.~V.; and Zoph, B. 2021{\natexlab{b}}.
\newblock Simple copy-paste is a strong data augmentation method for instance segmentation.
\newblock In \emph{Proceedings of the IEEE/CVF conference on computer vision and pattern recognition}, 2918--2928.

\bibitem[{Griffin et~al.(2018)Griffin, Caldwell, Andrews, and Bohler}]{5c62c410f56def97988aeeda}
Griffin, L.~D.; Caldwell, M.; Andrews, J. T.~A.; and Bohler, H. 2018.
\newblock “Unexpected Item in the Bagging Area”: Anomaly Detection in X-Ray Security Images.
\newblock 14: 1539--1553.

\bibitem[{Hsieh et~al.(2021)Hsieh, Robb, Chen, and Huang}]{hsieh2021droploss}
Hsieh, T.-I.; Robb, E.; Chen, H.-T.; and Huang, J.-B. 2021.
\newblock Droploss for long-tail instance segmentation.
\newblock In \emph{Proceedings of the AAAI conference on artificial intelligence}, volume~35, 1549--1557.

\bibitem[{Huang et~al.(2019)Huang, Wang, Chen, Xu, Tang, and Mu}]{huang2019modeling}
Huang, S.; Wang, X.; Chen, Y.; Xu, J.; Tang, T.; and Mu, B. 2019.
\newblock Modeling and quantitative analysis of X-ray transmission and backscatter imaging aimed at security inspection.
\newblock \emph{Optics express}, 27(2): 337--349.

\bibitem[{Ji, Shi, and Wang(2021)}]{ji2021prohibited}
Ji, Y.; Shi, C.; and Wang, X. 2021.
\newblock Prohibited item detection on heterogeneous risk graphs.
\newblock In \emph{Proceedings of the 30th ACM International Conference on Information \& Knowledge Management}, 3867--3877.

\bibitem[{Kang et~al.(2020)Kang, Xie, Rohrbach, Yan, Gordo, Feng, and Kalantidis}]{kang2020decoupling}
Kang, B.; Xie, S.; Rohrbach, M.; Yan, Z.; Gordo, A.; Feng, J.; and Kalantidis, Y. 2020.
\newblock Decoupling representation and classifier for long-tailed recognition.
\newblock \emph{Proceedings of the International Conference on Learning Representations}.

\bibitem[{Khan et~al.(2019)Khan, Jan, Farman, Ahmad, Farman, and Jan}]{khan2019deep}
Khan, M.; Jan, B.; Farman, H.; Ahmad, J.; Farman, H.; and Jan, Z. 2019.
\newblock Deep learning methods and applications.
\newblock \emph{Deep learning: convergence to big data analytics}, 31--42.

\bibitem[{LeCun, Bengio, and Hinton(2015)}]{lecun2015deep}
LeCun, Y.; Bengio, Y.; and Hinton, G. 2015.
\newblock Deep learning.
\newblock \emph{nature}, 521(7553): 436--444.

\bibitem[{Li, Liu, and Wang(2019)}]{li2019gradient}
Li, B.; Liu, Y.; and Wang, X. 2019.
\newblock Gradient harmonized single-stage detector.
\newblock In \emph{Proceedings of the AAAI conference on artificial intelligence}, volume~33, 8577--8584.

\bibitem[{Li et~al.(2024)Li, Zhikai, Lu, Lan, Cheung, and Huang}]{li2024feature}
Li, M.; Zhikai, H.; Lu, Y.; Lan, W.; Cheung, Y.-m.; and Huang, H. 2024.
\newblock Feature fusion from head to tail for long-tailed visual recognition.
\newblock In \emph{Proceedings of the AAAI conference on artificial intelligence}, volume~38, 13581--13589.

\bibitem[{Lu and Tong(2019)}]{lu2019towards}
Lu, J.; and Tong, K.-y. 2019.
\newblock Towards to Reasonable Decision Basis in Automatic Bone X-Ray Image Classification: A Weakly-Supervised Approach.
\newblock In \emph{Proceedings of the AAAI Conference on Artificial Intelligence}, volume~33, 9985--9986.

\bibitem[{Miao et~al.(2019)Miao, Xie, Wan, Su, Liu, Jiao, and Ye}]{miao2019sixray}
Miao, C.; Xie, L.; Wan, F.; Su, C.; Liu, H.; Jiao, J.; and Ye, Q. 2019.
\newblock Sixray: A large-scale security inspection x-ray benchmark for prohibited item discovery in overlapping images.
\newblock In \emph{Proceedings of the IEEE Conference on Computer Vision and Pattern Recognition}, 2119--2128.

\bibitem[{Mohan, Panas, and Cuadra(2020)}]{5cd9402ee1cd8e8e2dd1558f}
Mohan, K.~A.; Panas, R.~M.; and Cuadra, J.~A. 2020.
\newblock SABER: A Systems Approach to Blur Estimation and Reduction in X-ray Imaging.
\newblock 29: 7751--7764.

\bibitem[{Ren et~al.(2015)Ren, He, Girshick, and Sun}]{ren2015faster}
Ren, S.; He, K.; Girshick, R.; and Sun, J. 2015.
\newblock Faster r-cnn: Towards real-time object detection with region proposal networks.
\newblock In \emph{Advances in neural information processing systems}, 91--99.

\bibitem[{Ross and Doll{\'a}r(2017)}]{ross2017focal}
Ross, T.-Y.; and Doll{\'a}r, G. 2017.
\newblock Focal loss for dense object detection.
\newblock In \emph{proceedings of the IEEE conference on computer vision and pattern recognition}, 2980--2988.

\bibitem[{Sun et~al.(2021)Sun, Zhang, Jiang, Kong, Xu, Zhan, Tomizuka, Li, Yuan, Wang et~al.}]{sun2021sparse}
Sun, P.; Zhang, R.; Jiang, Y.; Kong, T.; Xu, C.; Zhan, W.; Tomizuka, M.; Li, L.; Yuan, Z.; Wang, C.; et~al. 2021.
\newblock Sparse r-cnn: End-to-end object detection with learnable proposals.
\newblock In \emph{Proceedings of the IEEE/CVF conference on computer vision and pattern recognition}, 14454--14463.

\bibitem[{Tan, Pang, and Le(2020)}]{tan2020efficientdet}
Tan, M.; Pang, R.; and Le, Q.~V. 2020.
\newblock Efficientdet: Scalable and efficient object detection.
\newblock In \emph{Proceedings of the IEEE/CVF conference on computer vision and pattern recognition}, 10781--10790.

\bibitem[{Tang, Huang, and Zhang(2020)}]{tang2020long}
Tang, K.; Huang, J.; and Zhang, H. 2020.
\newblock Long-tailed classification by keeping the good and removing the bad momentum causal effect.
\newblock \emph{Advances in neural information processing systems}, 33: 1513--1524.

\bibitem[{Tao et~al.(2022{\natexlab{a}})Tao, Li, Wang, Wei, Ding, Jin, Zhi, Liu, and Liu}]{EDS}
Tao, R.; Li, H.; Wang, T.; Wei, Y.; Ding, Y.; Jin, B.; Zhi, H.; Liu, X.; and Liu, A. 2022{\natexlab{a}}.
\newblock Exploring endogenous shift for cross-domain detection: A large-scale benchmark and perturbation suppression network.
\newblock In \emph{2022 IEEE/CVF Conference on Computer Vision and Pattern Recognition (CVPR)}, 21157--21167. IEEE.

\bibitem[{Tao et~al.(2022{\natexlab{b}})Tao, Wang, Wu, Liu, Liu, and Liu}]{FSOD}
Tao, R.; Wang, T.; Wu, Z.; Liu, C.; Liu, A.; and Liu, X. 2022{\natexlab{b}}.
\newblock Few-shot x-ray prohibited item detection: A benchmark and weak-feature enhancement network.
\newblock In \emph{Proceedings of the 30th ACM International Conference on Multimedia}, 2012--2020.

\bibitem[{Tao et~al.(2021)Tao, Wei, Jiang, Li, Qin, Wang, Ma, Zhang, and Liu}]{Hixray}
Tao, R.; Wei, Y.; Jiang, X.; Li, H.; Qin, H.; Wang, J.; Ma, Y.; Zhang, L.; and Liu, X. 2021.
\newblock Towards real-world X-ray security inspection: A high-quality benchmark and lateral inhibition module for prohibited items detection.
\newblock In \emph{Proceedings of the IEEE/CVF international conference on computer vision}, 10923--10932.

\bibitem[{Uijlings et~al.(2013)Uijlings, van~de Sande, Gevers, and Smeulders}]{5390b72e20f70186a0f21766}
Uijlings, J. R.~R.; van~de Sande, K. E.~A.; Gevers, T.; and Smeulders, A. W.~M. 2013.
\newblock Selective Search for Object Recognition.
\newblock 104: 154--171.

\bibitem[{Wagner et~al.(2020)Wagner, Cornet, Eckhoff, Andelfinger, Cai, and Knoll}]{wagner2020evaluation}
Wagner, M.; Cornet, H.; Eckhoff, D.; Andelfinger, P.; Cai, W.; and Knoll, A. 2020.
\newblock Evaluation of guidance systems at dynamic public transport hubs using crowd simulation.
\newblock In \emph{2020 Winter Simulation Conference (WSC)}, 123--134. IEEE.

\bibitem[{Wang et~al.(2021{\natexlab{a}})Wang, Zhang, Wen, Liu, and Wu}]{PIDray}
Wang, B.; Zhang, L.; Wen, L.; Liu, X.; and Wu, Y. 2021{\natexlab{a}}.
\newblock Towards real-world prohibited item detection: A large-scale x-ray benchmark.
\newblock In \emph{Proceedings of the IEEE/CVF international conference on computer vision}, 5412--5421.

\bibitem[{Wang et~al.(2021{\natexlab{b}})Wang, Zhang, Zang, Cao, Pang, Gong, Chen, Liu, Loy, and Lin}]{wang2021seesaw}
Wang, J.; Zhang, W.; Zang, Y.; Cao, Y.; Pang, J.; Gong, T.; Chen, K.; Liu, Z.; Loy, C.~C.; and Lin, D. 2021{\natexlab{b}}.
\newblock Seesaw loss for long-tailed instance segmentation.
\newblock In \emph{Proceedings of the IEEE/CVF conference on computer vision and pattern recognition}, 9695--9704.

\bibitem[{Wang et~al.(2025)Wang, Wang, Li, Ma, Wang, Jiang, and Tang}]{Wang_2025_CVPR}
Wang, X.; Wang, F.; Li, Y.; Ma, Q.; Wang, S.; Jiang, B.; and Tang, J. 2025.
\newblock CXPMRG-Bench: Pre-training and Benchmarking for X-ray Medical Report Generation on CheXpert Plus Dataset.
\newblock In \emph{Proceedings of the IEEE/CVF Conference on Computer Vision and Pattern Recognition (CVPR)}, 5123--5133.

\bibitem[{Wei et~al.(2020)Wei, Tao, Wu, Ma, Zhang, and Liu}]{WeiOccluded2020}
Wei, Y.; Tao, R.; Wu, Z.; Ma, Y.; Zhang, L.; and Liu, X. 2020.
\newblock Occluded Prohibited Items Detection: An X-Ray Security Inspection Benchmark and De-Occlusion Attention Module.
\newblock In \emph{Proceedings of the 28th ACM International Conference on Multimedia}, 138–146.

\bibitem[{Withers et~al.(2021)Withers, Bouman, Carmignato, Cnudde, Grimaldi, Hagen, Maire, Manley, Du~Plessis, and Stock}]{withers2021x}
Withers, P.~J.; Bouman, C.; Carmignato, S.; Cnudde, V.; Grimaldi, D.; Hagen, C.~K.; Maire, E.; Manley, M.; Du~Plessis, A.; and Stock, S.~R. 2021.
\newblock X-ray computed tomography.
\newblock \emph{Nature Reviews Methods Primers}, 1(1): 18.

\bibitem[{Xiao et~al.(2018)Xiao, Feng, Wei, Zhang, and Yan}]{xiao2018deep}
Xiao, H.; Feng, J.; Wei, Y.; Zhang, M.; and Yan, S. 2018.
\newblock Deep salient object detection with dense connections and distraction diagnosis.
\newblock \emph{IEEE Transactions on Multimedia}, 20(12): 3239--3251.

\bibitem[{Yang et~al.(2024)Yang, Jiang, Yan, Xue, Wang, and Wang}]{yang2024dual}
Yang, F.; Jiang, R.; Yan, Y.; Xue, J.-H.; Wang, B.; and Wang, H. 2024.
\newblock Dual-mode learning for multi-dataset X-ray security image detection.
\newblock \emph{IEEE Transactions on Information Forensics and Security}, 19: 3510--3524.

\bibitem[{Zhang(2017)}]{zhang2017mixup}
Zhang, H. 2017.
\newblock mixup: Beyond empirical risk minimization.
\newblock \emph{arXiv preprint arXiv:1710.09412}.

\bibitem[{Zhang, Chen, and Peng(2023)}]{zhang2023reconciling}
Zhang, S.; Chen, C.; and Peng, S. 2023.
\newblock Reconciling object-level and global-level objectives for long-tail detection.
\newblock In \emph{Proceedings of the IEEE/CVF International Conference on Computer Vision}, 18982--18992.

\bibitem[{Zhang et~al.(2020)Zhang, Chi, Yao, Lei, and Li}]{zhang2020bridging}
Zhang, S.; Chi, C.; Yao, Y.; Lei, Z.; and Li, S.~Z. 2020.
\newblock Bridging the gap between anchor-based and anchor-free detection via adaptive training sample selection.
\newblock In \emph{Proceedings of the IEEE/CVF conference on computer vision and pattern recognition}, 9759--9768.

\bibitem[{Zhang et~al.(2021{\natexlab{a}})Zhang, Wang, Cheng, Lu, Harrison, Xiao, Liao, and Miao}]{zhang2021window}
Zhang, X.; Wang, Y.; Cheng, C.-T.; Lu, L.; Harrison, A.~P.; Xiao, J.; Liao, C.-H.; and Miao, S. 2021{\natexlab{a}}.
\newblock Window loss for bone fracture detection and localization in x-ray images with point-based annotation.
\newblock In \emph{Proceedings of the AAAI Conference on Artificial Intelligence}, volume~35, 724--732.

\bibitem[{Zhang et~al.(2021{\natexlab{b}})Zhang, Wei, Zhou, and Wu}]{zhang2021bag}
Zhang, Y.; Wei, X.-S.; Zhou, B.; and Wu, J. 2021{\natexlab{b}}.
\newblock Bag of tricks for long-tailed visual recognition with deep convolutional neural networks.
\newblock In \emph{Proceedings of the AAAI conference on artificial intelligence}, volume~35, 3447--3455.

\bibitem[{Zhao et~al.(2022)Zhao, Zhu, Dou, Deng, and Wang}]{CLCXray}
Zhao, C.; Zhu, L.; Dou, S.; Deng, W.; and Wang, L. 2022.
\newblock Detecting overlapped objects in X-ray security imagery by a label-aware mechanism.
\newblock \emph{IEEE transactions on information forensics and security}, 17: 998--1009.

\bibitem[{Zhao, Teng, and Wang(2024)}]{zhao2024logit}
Zhao, L.; Teng, Y.; and Wang, L. 2024.
\newblock Logit normalization for long-tail object detection.
\newblock \emph{International Journal of Computer Vision}, 132(6): 2114--2134.

\bibitem[{Zhao et~al.(2019)Zhao, Zheng, Xu, and Wu}]{zhao2019object}
Zhao, Z.-Q.; Zheng, P.; Xu, S.-t.; and Wu, X. 2019.
\newblock Object detection with deep learning: A review.
\newblock \emph{IEEE transactions on neural networks and learning systems}, 30(11): 3212--3232.

\bibitem[{Zou et~al.(2023{\natexlab{a}})Zou, Chen, Shi, Guo, and Ye}]{zou2023object}
Zou, Z.; Chen, K.; Shi, Z.; Guo, Y.; and Ye, J. 2023{\natexlab{a}}.
\newblock Object detection in 20 years: A survey.
\newblock \emph{Proceedings of the IEEE}, 111(3): 257--276.

\bibitem[{Zou et~al.(2023{\natexlab{b}})Zou, Chen, Shi, Guo, and Ye}]{5cda948de1cd8ecf46bb4b94}
Zou, Z.; Chen, K.; Shi, Z.; Guo, Y.; and Ye, J. 2023{\natexlab{b}}.
\newblock Object Detection in 20 Years: A Survey.
\newblock 111: 257--276.

\end{thebibliography}



\end{document}